\documentclass[runningheads]{llncs}

\usepackage{eccv}

\usepackage{eccvabbrv}

\usepackage{graphicx}
\usepackage{booktabs}

\usepackage[accsupp]{axessibility} 
\usepackage[breaklinks,colorlinks,citecolor=eccvblue]{hyperref}

\usepackage{orcidlink}

\usepackage{xfrac}

\begin{document}

\title{Cyclic Data Parallelism for Efficient Parallelism of Deep Neural Networks} 

\titlerunning{Cyclic Data Parallelism for Efficient Parallelism of Deep Neural Networks}

\author{Louis Fournier\inst{1}
\and 
Edouard Oyallon\inst{2}%\thanks{This work was done partly before joining the Flatiron Institute.}
}

\authorrunning{L.~Fournier and E.~Oyallon}

\institute{ISIR, Sorbonne Université, CNRS, Paris, France \\\email{louis.fournier@isir.upmc.fr} \and
Center for Computational Mathematics, Flatiron Institute, New York 10010, USA} 

\maketitle

\begin{abstract}
  Training large deep learning models requires parallelization techniques to scale. In existing methods such as Data Parallelism or ZeRO-DP, micro-batches of data are processed in parallel, which creates two drawbacks: the total memory required to store the model's activations peaks at the end of the forward pass, and gradients must be simultaneously averaged at the end of the backpropagation step. We propose Cyclic Data Parallelism, a novel paradigm shifting the execution of the micro-batches from simultaneous to sequential, with a uniform delay. At the cost of a slight gradient delay, the total memory taken by activations is constant, and the gradient communications are balanced during the training step. With Model Parallelism, our technique reduces the number of GPUs needed, by sharing GPUs across micro-batches. Within the ZeRO-DP framework, our technique allows communication of the model states with point-to-point operations rather than a collective broadcast operation. We illustrate the strength of our approach on the CIFAR-10 and ImageNet datasets.
\end{abstract}

\section{Introduction}\label{intro}

Deep learning models have significantly increased in size in recent years, reaching hundreds of billions of parameters and requiring increasingly expensive training \cite{hoffmann2022training}. As the cost of training these models continues to rise, there is an increasing need for parallelization strategies. 
In particular, Data Parallelism (DP)~\cite{NIPS2012_6aca9700, 10.5555/2685048.2685095_scalingparameterserver} 
remains the dominant method for training Deep Neural Networks (DNNs) at scale.   
In DP, the model to be trained is first replicated on multiple workers. Then, during each training step, a mini-batch of data is sliced evenly among the workers in so-called micro-batches. Each worker then executes a forward and backward propagation on each micro-batch, and the locally computed gradients are subsequently averaged across all workers to obtain the gradient with regard to the full mini-batch. The resulting averaged gradient is used to locally update the models, typically following Stochastic Gradient Descent (SGD).

Nevertheless, DP has major drawbacks. First, the communication step between workers is synchronous, as all workers must complete their gradient computations before communicating, leading to idle workers waiting for the slowest worker~\cite{10.5555/2999611.2999748_stalesync}. Second, gradients are communicated globally with an all-reduce operation~\cite{goyal2018accurate}, which means that the communication step poses a challenge as the number of workers increases~\cite{collectivecom}. Last, the total memory used by all workers grows linearly with the number of workers since the model is fully replicated on each one~\cite{rajbhandari2020zero}. This requirement can be impractical since modern model sizes exceed the memory capacity of a single device (\eg, a GPU).

In this work, we tackle the memory and communication drawbacks of DP by proposing to change the execution time of workers in DP from simultaneous to sequential. We refer to this process as Cyclic Data Parallelism (CDP).
Our modification of standard DP aims to balance both the communication costs and the overall memory usage, by relying heavily on the sequential nature of the execution steps of DNNs during training.
More specifically, CDP reduces the cost of gradient communications in DP from collective communications at the end of the training step to point-to-point communications balanced over the entire training step. 
This balances the total memory used by all workers, at the cost of a small gradient delay. CDP can be combined with standard parallelization implementations for further improvements.

\paragraph{Contributions.}We list our specific contributions as follows:
\textbf{(a)} First, we propose CDP, an alternative to DP that balances gradient communications and overall memory usage across training, at the cost of computing on some delayed gradients.
    \textbf{(b)} We analyze the method by showing that it maintains a low communication and memory overhead during the computation of a mini-batch and allows a much more efficient implementation of mini-batch SGD on one or several GPUs. 
    \textbf{(c)} We then particularize the CDP paradigm to state-of-the-art approaches such as Model Parallelism (MP) and Zero Redundancy Optimizer powered DP (ZeRO-DP) \cite{rajbhandari2020zero}, showing improvements in all cases. 
    \textbf{(d)} Empirically, we show that the gradient delay of CDP leads to equal training of DNNs compared to DP on the large-scale CIFAR-10 and ImageNet datasets. 

\paragraph{Paper organization.}
Our paper is structured as follows. First, we summarize the parallelization techniques used to train DNNs in Sec. \ref{sec:related}. We then present standard mini-batch SGD with DP in Sec. \ref{sec:3-dp} before introducing our new CDP paradigm and two possible update rules in Sec. \ref{sec:3-cdp}. We then discuss how CDP improves on DP for the implementations of DP (on a single or multiple GPUs), MP and ZeRO-DP in Sec. \ref{sec:implems}. Finally, we present in Sec. \ref{sec:numerical} a numerical analysis of the update rules of CDP over the one of DP to train ResNets on CIFAR-10 and ImageNet and show the possible improvement in total memory for a ResNet-50 and a ViT-B/16.

\section{Related work}\label{sec:related}

Modern parallelization techniques for DNNs can be separated mainly into either DP, partitioning the mini-batch, or MP, partitioning the DNN model directly.

\paragraph{Data Parallelism.} 
In this class of techniques, the model is replicated on several workers, and each worker computes a micro-batch, a subset of the mini-batch. 
As noted above, this method suffers from a linear increase in memory as the number of workers increases, while communications costs increase either logarithmically or linearly~\cite{collectivecom}.
To reduce communications, less strict synchronization steps have been introduced, at the cost of updating the parameters with stale gradients~\cite{stich2019local, NIPS2011_218a0aef, 10.5555/2685048.2685095_scalingparameterserver}. However, these techniques do not take into account the memory required to train DNNs, a point we improve on.

Two notable DP techniques aim to reduce the memory cost of DP: Zero Redundancy Optimizer powered DP (\textit{ZeRO-DP})~\cite{rajbhandari2020zero} and \textit{Fully-Sharded DP}~\cite{zhao2023pytorch}. 
Rather than fully replicating the entire model across the DP workers, each worker retains only a subset of the model states (\ie optimizer states, gradients, and parameters) corresponding to one stage of the model. 
The memory required can be further reduced by storing model states on the CPU~\cite{ren2021zerooffload, rajbhandari2021zeroinfinity}, at the expense of communications between the CPU and the GPUs. Despite the memory gains, the volume of communication increases in ZeRO-DP, which has been addressed by compressing communications or striking a balance with the memory~\cite{zhang2022mics, wang2023zero}.  
Nevertheless, the simultaneous nature of the micro-batches executions remains the root cause of ZeRO-DP's issues, which we tackle in this paper.

\paragraph{Model Parallelism.} Instead of parallelizing computations by slicing a mini-batch, MP~\cite{NIPS2012_6aca9700, shazeer2018meshtensorflow} aims to partition the components of the DNN.  
MP can be separated into two categories. 
First, Intra-layer MP~\cite{jia2019beyond, shoeybi2019megatron}, also called tensor parallelism, partitions individual layers across workers but requires a high communication cost. 
Second, Inter-layer MP~\cite{NIPS2012_6aca9700} splits a DNN into successive stages, and workers are solicited for only one stage, thus requiring less memory. 
A disadvantage of this method is that only one of the workers computes at a time since the stages are executed sequentially in the forward-backward pass. 
This limitation was addressed in several ways. Local learning~\cite{nøkland2019training, belilovsky2020decoupled, pmlr-v202-fournier23a} 
removes the need for a backward pass of the model with local auxiliary objectives at each stage. This allows stages to compute mini-batches simultaneously but at the expense of model performance. Delayed gradient methods~\cite{pmlr-v139-zhuang21a_accumulated, xu2022acceleration} allow stages to compute in parallel on micro-batches, by allowing updates with delayed gradients. 

Of particular interest is \textit{Pipeline Parallelism} (PP), a form of Inter-layer MP, introduced in GPipe~\cite{huang2019gpipe}, which also avoids this problem.
In PP, mini-batches are divided into micro-batches which are passed sequentially to the successive stages.
This allows each micro-batch to be propagated to the next stage as soon as it has completed its forward pass, allowing stages to compute micro-batches in parallel. GPipe thus reduces the number of idle workers, while still leaving a "bubble". 
PipeDream~\cite{narayanan2019pipedream} proposes to reduce further reduce this bubble of computation by introducing the one-forward-one-backward (1F1B) pipeline schedule. Within this implementation, a stage alternates between the forward and the backward pass of one micro-batch. However, this comes at a cost as multiple versions of the parameters must be stored by a stage, and potentially storing the activations of the entire batch. Our method similarly introduces a cyclic delay of micro-batches. However, our framework is more general and notably applies to many other parallelism frameworks, not just PP. GEMS\cite{9355254_gems} is a MP framework that similarly reduces worker idleness by offsetting computations, like CDP and PP. Still, workers must store parameters of two stages, and GEMS does not extend to other frameworks either, unlike CDP. 
PipeDream-2BW~\cite{narayanan2021memory} 
improves on PipeDream by keeping only two versions of the parameters and accumulating the gradients of the micro-batches to update them. This allows for continuous computations on each stage with no idle workers at the cost of a gradient staleness of one mini-batch. We will show that our more general framework can recover and improve on the update rule of PipeDream-2BW when applied to the particular case of PP. 
PipeMare~\cite{yang2021pipemare} keeps a single version of the weights and, like the previous delayed gradient methods, allows asynchronous updates. However, it requires additional hyperparameters, which we do not. DAPPLE~\cite{fan2021dapple} improves on PP by using stage replicas to handle micro-batches partitions. This can be combined with our CDP framework. Still, as in PP, GPUs need to store batch activations, not just micro-batches. Finally, note that the parallelization techniques we have discussed are often integrated together, with 3D Parallelism~\cite{smith2022using} being a notable example that combines DP, tensor parallelism, and PP.

\begin{figure*}%[h]
    \centering
    \includegraphics[width=0.99\linewidth]{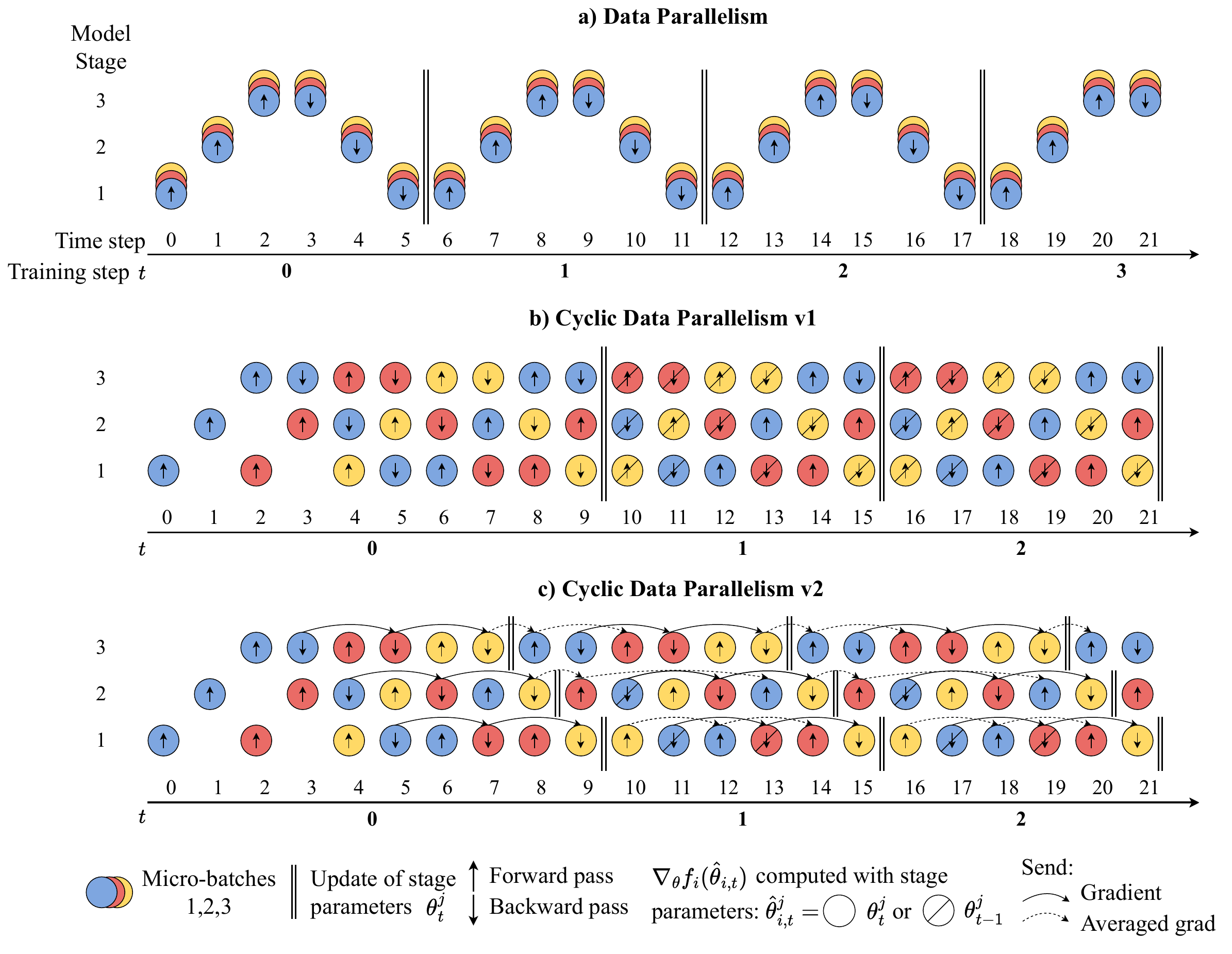}
\begin{subfigure}{0\linewidth}
\phantomsubcaption\label{fig:timeline-dp}
\end{subfigure}% <----- get rid of space, for proper centering
\begin{subfigure}{0\linewidth}
\phantomsubcaption\label{fig:timeline-cdp1}
\end{subfigure}% <----- get rid of space, for proper centering
\begin{subfigure}{0\linewidth}
\phantomsubcaption\label{fig:timeline-cdp2}
\end{subfigure}% <----- get rid of space, for proper centering
    \caption{\textbf{Timeline of executions for Data Parallelism (DP) and the two versions of Cyclic Data Parallelism (CDP), for $N{=}3$ workers.} \textbf{(a) DP.} The 3 workers begin executing their forward pass simultaneously in DP, and maintain this synchronization throughout the entire forward-backward pass. \textbf{(b) CDP-v1.} In CDP, the 3 workers begin executing with an equal delay between them (equal to $2$ time steps). For CDP-v1 (see Eq.~\eqref{eq:ur_tm1}), the parameters of the model are updated with a delay constant equal to one training step. \textbf{(c) CDP-v2.} This delay is limited in CDP-v2 (see  Eq.~\eqref{eq:ur_int}), by allowing the stages to update and send gradients independently. The communication scheme, balanced across the training step, is indicated. Note that the total complexity of a training step (a forward-backward pass) does not change, but activation memory does not peak in CDP as it does in DP.}
    \label{fig:timeline}
\end{figure*}

\section{The Cyclic Data Parallelism paradigm}

\subsection{A motivating example: mini-batch SGD with Data Parallelism}\label{sec:3-dp}
The Cyclic Data Parallelism paradigm is best understood from the perspective of the standard DP paradigm. We first describe it before highlighting its caveats. 

\paragraph{Mini-batch SGD via DP.}  A standard DP strategy applied to a mini-batch SGD step is to replicate a model over $N$ workers. Each worker is then fed with mini-batch slices, referred to as micro-batches~\cite{huang2019gpipe}. 
At training step $t$, each replica $i\in[1,N]$ receives a micro-batch of data leading to a training objective $f_i$ parametrized by $\theta_t$, and computes its corresponding gradient $\nabla f_i(\theta_t)$. We assume that the model can be partitioned into $N$ stages, and write $\theta_t=\{\theta^j_t\}_{j\in [1,N]}$ to emphasize the dependency in the stage $j$. Parameter gradients are computed via a forward and backward pass over the micro-batch. They are averaged over all workers, typically using an all-reduce operation \cite{collectivecom}. Finally, each worker locally updates the parameters $\theta_t$ with the averaged gradient. 
With the learning rate $\gamma_t$, the standard update rule~\cite{Robbins1951ASA} which results 
can be written as
\begin{equation}\label{eq:ur_t}\tag{DP}
    \theta_{t+1} = \theta_t - \frac{\gamma_t}{N} \sum_{i=1}^N \nabla  f_i(\theta_t).
\end{equation}

\paragraph{Disadvantages of DP.} The execution timeline of DP is illustrated in Fig. \ref{fig:timeline-dp}, where one time step corresponds to the execution of a forward or backward pass of a stage. We assume that the forward and backward passes of a stage take a similar amount of time for the $N$ workers. A training step $t$ (which indexes the sequence of the parameters) is thus composed of $2N$ time steps. 
There are several inherent problems with the DP strategy applied to mini-batch SGD. First, the \textbf{total number of retained activations} peaks every $2N$ time steps, notably occurring at step $2$ in Fig. \ref{fig:timeline-dp}. This peak results from the intermediate activations that are stored and await release during the backward pass. While this memory overhead can be mitigated by strategies such as pipelining~\cite{huang2019gpipe}, there exists an unavoidable waiting barrier every $2N$ steps to synchronize all gradient computations. This notably occurs between steps 5 and 6 in Fig. \ref{fig:timeline-dp}. Consequently, this barrier introduces \textbf{latency} and leads to \textbf{workers idling}. Furthermore, this waiting barrier also leads to a \textbf{communication overhead}, as workers must simultaneously communicate their local buffers. While this can be solved by a collective all-reduce operation~\cite{collectivecom}, communication can be an issue in standard centralized frameworks, such as Federated Learning \cite{Konecn2016FederatedLS}. 

\subsection{Towards delayed mini-batch SGD with Cyclic Data Parallelism}\label{sec:3-cdp}

We now propose a method that avoids these problems, caused by the simultaneous execution of computational steps in DP. 
\paragraph{Algorithmic description.} Contrary to the previous DP approaches, our main idea is to break the synchrony between the forward and backward passes of the $N$ micro-batches. Instead of each step being computed simultaneously, we assume that every micro-batch computation is delayed by an identical number of time steps (\eg, if one worker handles one micro-batch, it will start with a delay). 
More specifically, the computations of $\nabla f_i$ are delayed by a delay of $2$ time steps compared to $\nabla f_{i-1}$ (where $i$ is taken modulo $N$, with $N$ the number of stages and micro-batches like before). 
This results in a cyclic pattern, illustrated in Fig. \ref{fig:timeline-cdp1} and \ref{fig:timeline-cdp2}. 
Furthermore, each stage constantly performs either a forward or a backward pass on a single and distinct micro-batch: 
in particular, it implies that the maximum amount of activations stored at a given time step is nearly constant during training, and in this case, smaller compared to DP. We call this concept \textbf{Cyclic Data Parallelism} (CDP). 

\paragraph{Update rules.} While the forward and backward passes are fully defined assuming an available parameter $\theta_t$, one has to define the corresponding update rule to produce $\theta_{t+1}$, since breaking the synchrony between the backward passes of the micro-batches makes it impossible to follow Eq.~\eqref{eq:ur_t}.  Similarly to \cite{chen2018efficient, yang2021pipemare}, at the training step $t$ we introduce an auxiliary variable $\{\hat \theta_{i,t}\}_i$, a buffer updated by other concurrent workers. Formally, our algorithm can be written as
\begin{align}\label{eq:ur_cdp}\tag{CDP}
    \theta_{t+1} &= \theta_t - \frac{\gamma_t}{N} \sum_{i=1}^N \nabla  f_i(\hat \theta_{i,t})\,,\\ 
    \hat \theta^j_{i,t} &= u_{i,j}(\theta^j_t,\theta^j_{t-1})\,. \nonumber
\end{align}
where $u_{i,j}(a,b)\in\{a,b\}$ 
is a rule on the parameters, that depends on both the micro-batch $i$ and the stage $j$. 
Thus, $u_{i,j}$ implicitly depends on the time step of the algorithm but not on the training step in our paradigm. An alternative that would is possible,  and would easily allow more complex asynchronous or randomized variants. Note that some rules $u_{i,j}$ are not possible due to the delay between workers, preventing for instance the update rule of DP. 

We propose two update rules in particular that follow from Eq.~\eqref{eq:ur_cdp} with specific rules $u_{i,j}$. They can be considered as the edge cases of the update rule, with maximum and minimum delay respectively, with all other rules $u_{i,j}$ being an intermediary between them. First, we use a simultaneous barrier after the $N$th batch finishes its computation, which is illustrated in Fig. \ref{fig:timeline-cdp1} with a barrier at the time step $9$. In this specific case, 
a consistent update consists of the rule $u_{i,j}(a,b)=b$, using the stored $\hat \theta_{i,t}=\theta_{t-1}$. 
This leads to the delayed update rule
\begin{align}\label{eq:ur_tm1}
    \theta_{t+1} &= \theta_t - \frac{\gamma_t}{N} \sum_{i=1}^N \nabla  f_i( \theta_{t-1}). \tag{CDP-v1}
\end{align}

We refer to this update rule as CDP-v1. In the specific case of PP, we recover the update rule of PipeDream-2BW \cite{narayanan2021memory}. We now discuss another update rule, which we refer to as CDP-v2, and which is illustrated in Fig. \ref{fig:timeline-cdp2}. We fix the 
 rule $u_{i,j}$ such that after the $N$th micro-batch finishes the gradient computation of a stage, it directly transmits the updated stage parameters to be computed on a micro-batch (the order of communication can be predefined). This gives the rule $u_{i,j}(a,b)=a$ if $j \geq N-i+1$, and $b$ otherwise, reducing the number of delayed parameters compared to CDP-v1. Note also that we do not need to keep a copy of two parameters, as the parameters in memory are always the freshest available. 
This update is written as 
\begin{equation}\label{eq:ur_int}
    \theta_{t+1} = \theta_t - \frac{\gamma_t}{N} \sum_{i=1}^N \nabla  f_i(\theta^1_{t-1}, ..., \theta^{N-i}_{t-1}, \theta^{N-i+1}_{t}, ..., \theta^N_{t})\,.\tag{CDP-v2}
\end{equation}
Note that in both cases, gradient communications need not occur \emph{simultaneously}, instead, results can be communicated at intermediate steps, benefiting the overall communication bandwidth. We propose a possible communication scheme for CDP-v2 in Fig. \ref{fig:timeline-cdp2}.

\paragraph{Remark on the convergence.} We emphasize that the generic update of Eq.~\eqref{eq:ur_cdp}, in the worst case, 
can be understood as SGD with delayed gradients with a fixed delay of 1. There is a wide body of literature that studies the convergence of delayed first-order methods~\cite{mishchenko2023asynchronous, JMLR:v21:19-748_stich_sgd_delay, pmlr-v162-yang22r}, which guarantee that our rate of convergence is almost equal to the one of SGD. In practice, very large DNNs have been shown to converge similarly with or without a delay of 1~\cite{narayanan2021memory, ren2021zerooffload}, and models up to 530B parameters converge with larger delays with no issues reported~\cite{shoeybi2019megatron}.  

\begin{figure*}[ht]
    \centering
    \includegraphics[width=0.8\linewidth]{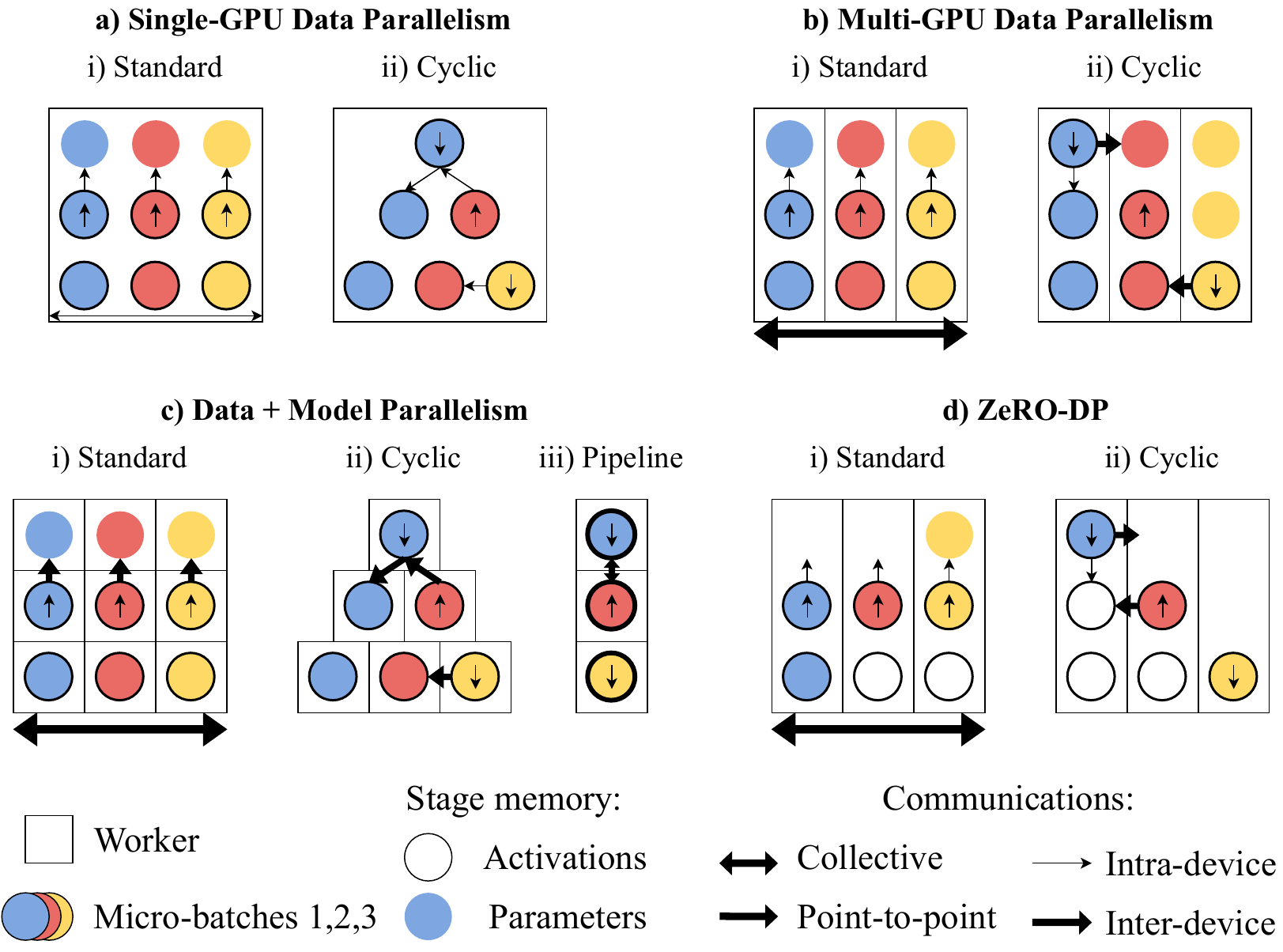}
    \caption{\textbf{Comparison between parallelism frameworks with and without using CDP}, for $N{=}3$. A device (\eg, a GPU) is represented by a rectangle, and the different micro-batches being computed by the $3$ colors. A model stage requires memory, for the parameters used for computation and for the activations retained awaiting the backward pass, indicated with a colored disk and a black circle. 
    Communications are intra or inter-device (thin or thick arrow), collective or point-to-point (double-headed or single-headed arrow). \textbf{(a) Single-GPU DP.} This setting corresponds to a high-connectivity device with limited memory. We observe a memory reduction of half. \textbf{(b) Multi-GPU DP.} Communications can be drastically reduced when using multiple GPUs with CDP. \textbf{(c) DP+MP.} Both the number of required GPUs and the communications are reduced compared to a standard implementation of MP with DP. 
    Only $N$ GPUs are needed in PP, but they require more activation memory, shown with a thicker circle. 
    \textbf{(d) ZeRO-DP.} The model states needs to be sent or received by only one worker at each time step, instead of the standard broadcast operation of ZeRO-DP.}
    \label{fig:comparison}
% this awful code allows for the specifics labels
%\renewcommand{\p@subfigure}{\thefigure}
\renewcommand{\thefigure}{2.a.}
\renewcommand{\thesubfigure}{\roman{subfigure}}
\begin{subfigure}{0\linewidth}
\phantomsubcaption\label{fig:comp-ai}
\end{subfigure}% <----- get rid of space, for proper centering
\begin{subfigure}{0\linewidth}
\phantomsubcaption\label{fig:comp-aii}
\end{subfigure}
%\counterwithin{subfigure}{section}
\renewcommand{\thefigure}{2.b.}
\renewcommand{\thesubfigure}{i.}
\begin{subfigure}{0pt}
\phantomsubcaption\label{fig:comp-bi}
\end{subfigure}% <----- get rid of space, for proper centering
\renewcommand{\thefigure}{2.b.}
\renewcommand{\thesubfigure}{ii.}
\begin{subfigure}{0pt}
\phantomsubcaption\label{fig:comp-bii}
\end{subfigure}% <----- get rid of space, for proper centering
\renewcommand{\thefigure}{2.c.}
\renewcommand{\thesubfigure}{i.}
\begin{subfigure}{0\linewidth}
\phantomsubcaption\label{fig:comp-ci}
\end{subfigure}% <----- get rid of space, for proper centering
\renewcommand{\thefigure}{2.c.}
\renewcommand{\thesubfigure}{ii.}
\begin{subfigure}{0pt}
\phantomsubcaption\label{fig:comp-cii}
\end{subfigure}% <----- get rid of space, for proper centering
\renewcommand{\thefigure}{2.c.}
\renewcommand{\thesubfigure}{iii.}
\begin{subfigure}{0pt}
\phantomsubcaption\label{fig:comp-ciii}
\end{subfigure}% <----- get rid of space, for proper centering
\renewcommand{\thefigure}{2.d.}
\renewcommand{\thesubfigure}{i.}
\begin{subfigure}{0pt}
\phantomsubcaption\label{fig:comp-di}
\end{subfigure}% <----- get rid of space, for proper centering
\renewcommand{\thefigure}{2.d.}
\renewcommand{\thesubfigure}{ii.}
\begin{subfigure}{0pt}
\phantomsubcaption\label{fig:comp-dii}
\end{subfigure}% <----- get rid of space, for proper centering
\end{figure*}

\section{Variants of DP, MP and Zero-DP with CDP}\label{sec:implems}

We now discuss how CDP reduces the computational overhead of DP (on one or several GPUs), MP, and ZeRO-DP. A schematic representation of the implementations is presented in Fig. \ref{fig:comparison}, and the theoretical values of the communication and memory requirements are summarized in Tab. \ref{tab:costs}, which we discuss next.

\begin{table*}%\edouard{
    \centering
    \resizebox{\columnwidth}{!}{
    \begin{tabular}{l|llllll}
         & \multicolumn{2}{c}{Memory per GPU}  & \multicolumn{2}{c}{Communications inter-GPUs} &   &  \\ %\hline  
         Implementation & Activations & Parameters & Volume & Max com. steps & Nb of GPUs & Rule \\ \hline  
         Single-GPU DP & $NB\Psi_A$ &  $N\Psi_P$ &  &  & 1 & \eqref{eq:ur_t} \\
        + Cyclic & \boldmath$\frac{(N{+}1)}{2} B\Psi_A$ & \boldmath$\frac{(N{+}1)}{2}\Psi_P$ &   &   & 1 &  \eqref{eq:ur_cdp} \\ \hline
        Multi-GPU DP & $B\Psi_A$ & $\Psi_P$ & $\Psi_P$ & $\mathcal{O}(\log(N))$ & $N$ & \eqref{eq:ur_t} \\
        + Cyclic & $B\Psi_A$ & $\Psi_P$ & $\Psi_P$ & \boldmath$\mathcal{O}(1)$ & $N$ &  \eqref{eq:ur_cdp} \\ \hline
        DP with MP & $B \Psi_A/N$ & $\Psi_P / N$ & $\Psi_P{+}B\Psi^{\text{int}}_{A}$ & $\mathcal{O}(\log(N))$ & $N^2$ &  \eqref{eq:ur_t}  \\ 
        + Cyclic & $B \Psi_A/N$ & $\Psi_P / N$ & \boldmath$\frac{1}{2}\Psi_P{+}B\Psi^{\text{int}}_{A}$ & \boldmath$\mathcal{O}(1)$ & \boldmath$\frac{1}{2}(N{+}1)N$ & \eqref{eq:ur_cdp} \\ %\hline
        PP & $B\Psi_A$ & $\Psi_P / N$ & $B\Psi^{\text{int}}_{A}$ & $\mathcal{O}(1)$ & $N$ & \eqref{eq:ur_cdp}  \\  \hline
        ZeRO-DP & $B\Psi_A$ & $\Psi_P / N$ & $\Psi_P$ & $\mathcal{O}(\log(N))$ & $N$ & \eqref{eq:ur_t} \\
        + Cyclic &$B\Psi_A$ & $\Psi_P / N$ & $\Psi_P$ & \boldmath$\mathcal{O}(1)$ & $N$ & \eqref{eq:ur_cdp} \\ %\hline
    \end{tabular}
    }
    \caption{\textbf{Theoretical cost of the parallelism implementations discussed in Sec. \ref{sec:implems}.} All implementations are improved by using CDP (Cyclic) over DP, in terms of memory (per GPU or number of GPUs) or communication. Improvements are noted in bold. The parameter memory of the entire model is noted as $\Psi_P$, and the activation memory of the entire model for one data sample by $\Psi_A$, or $\Psi^{\text{int}}_{A}$ for the subset communicated in MP. 
    $N$ indicates both the number of stages and of micro-batches (of size $B$). The communication volume is the size of the tensors that need to be communicated. `Max com. steps' is the maximum number of steps of communication required between $2$ time steps between two workers, which is a minimum of $\mathcal{O}(\log(N))$ steps for a collective operation or a single step for point-to-point communications. The update rule used is specified. Note that the parameter memory required in Single-GPU DP depends on the implementation.}
    \label{tab:costs}
\end{table*}

\paragraph{Analytical comparisons.} 
We will refer to $\Psi_P$ as the memory occupied by the parameters of the entire model (including the optimizer states here). $\Psi_A$ refers to the memory occupied by the activations of one data sample in the entire model. In MP, stages communicate activations at the inputs and outputs of stages. This subset of activations ouccupies a memory that we note $\Psi^{\text{int}}_{A}$ (and thus $\Psi^{\text{int}}_{A}\leq\Psi_{A}$). From now on, $N$ refers to the number of stages of the model as well as the number of micro-batches, which have equal size $B$. 
In Tab. \ref{tab:costs}, we summarize the memory requirements per device, the communication volume, and the maximum number of communication steps required between $2$ time steps, for DP, MP, PP, and ZeRO-DP with and without CDP. We find the following improvements with CDP compared to DP. CDP halves the memory required to train on DP with a single-GPU device. Multi-GPU DP with CDP doesn't require an all-reduce operation at the end of the training step but only point-to-point communications at each time step. This improvement in the number of communications required between time steps is also found with MP and ZeRO-DP with CDP. CDP allows MP to halve the number of GPUs needed, thus halving the memory required, as well as the communication of gradients between GPUs. The number of GPUs can be further reduced from MP with CDP to PP, which can be seen as a particular implementation of CDP. However, this requires the GPUs to be able to store the activations of the entire mini-batch. More generally, a disadvantage of MP and PP is that they need to communicate activations, which scale with the batch size $B$. We will now discuss each setting in more depth.

\subsection{CDP implementation on a single-GPU device}
We first explore the setting of a single-GPU device training with mini-batch SGD, assuming that communication within the GPU is cheap and fast, but memory is limited~\cite{9361255}.

\paragraph{Standard implementation.} 
We illustrate the training of standard DP on a single-GPU device in Fig. \ref{fig:comp-ai}. Then, at the peak of the forward pass, the GPU must retain the activations of the entire model for $N$ micro-batches, representing the total mini-batch, resulting in a total memory of activations of about $NB \Psi_A$.

\paragraph{Cyclic implementation.} Turning our attention to CDP, depicted in Fig. \ref{fig:comp-aii}, we find that its implementation on a single-GPU device requires about half the memory compared to standard DP. Indeed, in this approach, each stage of the model processes one micro-batch at each time step. Consequently, the total memory occupied by activations remains nearly constant across time steps, equal approximately to $\frac{(N{+}1)}{2}B \Psi_A $, reducing in half the total memory required for a DP implementation. Depending on the implementation, parameters may be shared in the GPU, resulting in no parameter memory improvement. The memory improvement is still significant in this case, as activation memory is generally much larger than parameter memory, peaking at 60GB for a GPT-2 model in \cite{rajbhandari2020zero}, compared to 3GB for the fp16 parameters. 

\subsection{CDP implementation on a multi-GPU device}
Here, one GPU processes a single micro-batch, but communications between GPUs may hinder the overall system efficiency. 

\paragraph{Standard implementation.} We consider the case where DP is implemented on $N$ GPUs that process $N$ micro-batches of data independently, as depicted in Fig. \ref{fig:comp-bi}. These GPUs communicate the micro-batches gradients with an all-reduce operation at the end of the training step before proceeding to the next time step. All-reduce is a collective operation that requires at least $\mathcal{O}(\log(N))$ communications steps \cite{collectivecom} in a favorable setting, or $\mathcal{O}(N)$ steps for a bandwidth-optimal ring-based implementation \cite{PATARASUK2009117_ringreduce}. 

\paragraph{Cyclic implementation.} CDP on $N$ GPUs, as depicted in Fig. \ref{fig:comp-bii}, makes better use of the communication bandwidth between GPUs, as the all-reduce operation is fragmented across the training step into point-to-point operations between each time step. Indeed, half of the stages compute a backward pass at any time step, and then send the computed gradient to the following worker (modulo $N$) for the reduce operation. This communication scheme corresponds to the ring-based all-reduce operation \cite{PATARASUK2009117_ringreduce}, which has optimal bandwidth usage.

\subsection{Implementation in a MP paradigm}\label{sssec:mp}

\paragraph{Standard implementation.}
Now, in MP, a GPU does not hold a full replica of a model but only of a single stage. During the forward and backward passes, activations and gradients must be communicated between GPUs holding successive stages. 
A naive implementation of DP and MP results in Fig. \ref{fig:comp-ci}, where the $N$ micro-batches require $N^2$ GPUs to process each slice of the data and the model. Furthermore, note also that only $N$ GPUs are busy at a time, thus most GPUs are idle, waiting for the forward pass of the next batch of data which is a significant inefficiency of MP. 

\paragraph{Cyclic implementation.} Here, CDP with MP, which we present in Fig. \ref{fig:comp-cii}, once again improves on the DP implementation. In particular, the number of GPUs required for CDP is reduced, and we can show that for each stage $j$, CDP requires only $N-j+1$ GPUs to be shared among the $N$ micro-batches. This leads to a "pyramidal" shape of the stage structure. Thus in total, only $\frac 12N(N{+}1)$ GPUs holding a stage of the model are needed, halving the number over DP, as for the memory required. Compared to DP, a micro-batch doesn't send the activations of a stage to the same GPU every time for the computation of the next stage. Instead, a GPU in the previous time step will have released the activations stored in its memory after a backward pass. The micro-batch will send the activations to this GPU as its memory is available for computations. 
Since the number of devices is smaller in DP than in CDP, note also that the total number of gradient communications between devices is reduced. 
In fact, if a GPU is only able to hold the activation of one micro-batch, the number of GPUs used, $\frac 12N(N{+}1)$, is optimal to compute $N$ micro-batches. An in-depth discussion is proposed in the Appendix of this paper.

\paragraph{Pipeline Parallelism implementation.} If we assume that one GPU can store the activations of all $N$ micro-batches, then we can further reduce the total number of devices needed to only $N$, one per stage. Indeed, this implementation of CDP with MP on $N$ devices, depicted in Fig. \ref{fig:comp-ciii}, is equivalent to PP as presented in PipeDream \cite{narayanan2019pipedream}. If we follow our first update rule Eq.~\eqref{eq:ur_tm1}, CDP follows the same update rule as PipeDream-2BW \cite{narayanan2021memory}. However, our second update rule Eq.~\eqref{eq:ur_int} improves on this update rule by reducing the gradient delay. 
PP requires fewer GPUs than MP, but only if the GPUs can store the activations of the entire mini-batch. This limits its scaling capacity compared to MP, which is $2$ times more GPU-efficient. Indeed, for GPUs with similar memory capacities, MP with CDP requires $\frac{1}{2}(N+1)N$ GPUs to train a batch of size $B$. Meanwhile, PP requires $N$ GPUs to train a batch of size $B/N$, or $N^2$ GPUs to train a batch of size $B$ by combining PP with DP.

\subsection{Implementation in a ZeRO-DP paradigm} \label{sssec:zero}
\paragraph{Standard implementation.} ZeRO-DP \cite{rajbhandari2020zero} is a training framework that aims at combining the advantages of DP and MP, which we represent in Fig. \ref{fig:comp-di}. Rather than replicating the entire model on every one of the $N$ GPUs, ZeRO-DP replicates only the model states of a single stage across the GPUs. The model state refers (in stage 3 of ZeRO-DP) to the model parameters, gradients, and optimizer states. When the workers execute the forward or backward propagation through a stage, the model states of that stage are broadcast from the GPU that stores them to all GPUs. After computation, the model states are deleted to free up the memory, such that a GPU only retains the model states of a maximum of two stages. The communication volume is similar to standard multi-GPU DP, with a $1.5$ increase at most. However note that like in PP, the memory taken by the activations on one GPU increases with $N$ since all stages of the model must be retained.
\paragraph{Cyclic implementation.} CDP improves on ZeRO-DP by removing the need for collective communications of the model states, as depicted in Fig.~\ref{fig:comp-dii}. Since one stage is computed by a single GPU at a time step, the model states only need to be held by this GPU, with no replication. Then, they only need to be communicated to a single GPU at the next time step. 
This reduces the communication overhead at each time step, similarly to Multi-GPU DP.

\section{Numerical analysis}\label{sec:numerical}

\begin{table}[t]
    \begin{subtable}[h]{0.45\textwidth}
    \centering
    \begin{tabular}{c|ccc}
    %\hline
         & \multicolumn{3}{c}{Learning Rule} \\
        %\hline
         Model& \eqref{eq:ur_tm1} & \eqref{eq:ur_int}& \eqref{eq:ur_t}  \\\hline
        ResNet-18 & 94.1  &\textbf{94.8}& 94.7 \\%\hline
        ResNet-50 & 94.0& \textbf{94.5}  &  \textbf{94.5} \\%\hline
    \end{tabular}
        \caption{Test accuracy on CIFAR10}
    %\caption{}
    \label{tab:xp_rules_cifar10}
    \end{subtable}
    \hfill
    \begin{subtable}[h]{0.45\textwidth}
    \centering
    \begin{tabular}{c|ccc}
    %\hline
         & \multicolumn{3}{c}{Learning Rule} \\
        %\hline
         Model& \eqref{eq:ur_tm1} & \eqref{eq:ur_int}& \eqref{eq:ur_t}  \\\hline
         ResNet-18 & 69.9 & 70.0  & \textbf{70.1}  \\ %\hline
         ResNet-50 & \textbf{75.8} & 75.7  & 75.4\\ 
    \end{tabular}
        \caption{Test accuracy on ImageNet}
    \label{tab:xp_rules_imagenet}
        \end{subtable}
        \caption{\textbf{Top-1 test accuracy for the three learning rules \eqref{eq:ur_tm1}, \eqref{eq:ur_int} and \eqref{eq:ur_t} on the (a) CIFAR10 and (b) ImageNet datasets}, by training a ResNet-18 and a ResNet-50. Our results are stable, as the standard deviation over 5 runs is systematically less than 0.08. We observe that on CIFAR10, CDP systematically performs similarly or better than DP, especially CDP-v2. On ImageNet, CDP performs similar to or better than DP. }
\end{table}

\paragraph{Hyperparameters.} To test our method, we propose to use the standard training pipeline on CIFAR-10 and ImageNet datasets, trained on the ResNet-18 and ResNet-50 architectures, split into 4 stages with similar FLOPs. We simulate this partitioning using the \texttt{fvcore} library \footnote{\hyperlink{https://github.com/facebookresearch/fvcore/}{https://github.com/facebookresearch/fvcore/}} to compute the FLOPs count of each ResNet's module. For a finer partition of the linear modules, we separate them into weight and bias modules, by approximating the FLOPs of the bias module as the square root of the linear module's FLOPs. 
We consider the SGD optimizer with momentum 0.9 and we simulate our delayed activations for DP, CDP-v1, and CDP-v2. We train over 100 epochs with batch size 128 and 90 epochs with batch size 256 respectively for CIFAR-10 and ImageNet. 
We consider an initial learning rate of 0.05 and 0.1, dropped by a factor of 0.2 and 0.1 on CIFAR-10 and ImageNet respectively, at epochs 30, 60, and 90. The weight decay is equal to $10^{-4}$ for ImageNet, and $5{\times}10^{-4}$ or $10^{-3}$ for CIFAR-10, whether trained on a ResNet-18 or 50. To accommodate the smaller image size of CIFAR-10, as standard, we remove the first max pooling and reduce the kernel size of the first convolutional layer to 3 and the stride to 1. 
For CIFAR-10, we report the accuracy and its corresponding standard deviation on the test sets for 5 runs. For ImageNet, following standard practice, we report the maximum validation accuracy over the last 10 epochs. Contrary to \cite{yang2021pipemare}, we did not need any specific tuning of the other hyper-parameters for our update rules. Our source code is available at   \hyperlink{https://github.com/fournierlouis/Cyclic\_Data\_Parallelism}{github.com/fournierlouis/Cyclic\_Data\_Parallelism}.

\begin{figure}[ht]
    \centering
    \includegraphics[width=0.4\linewidth]{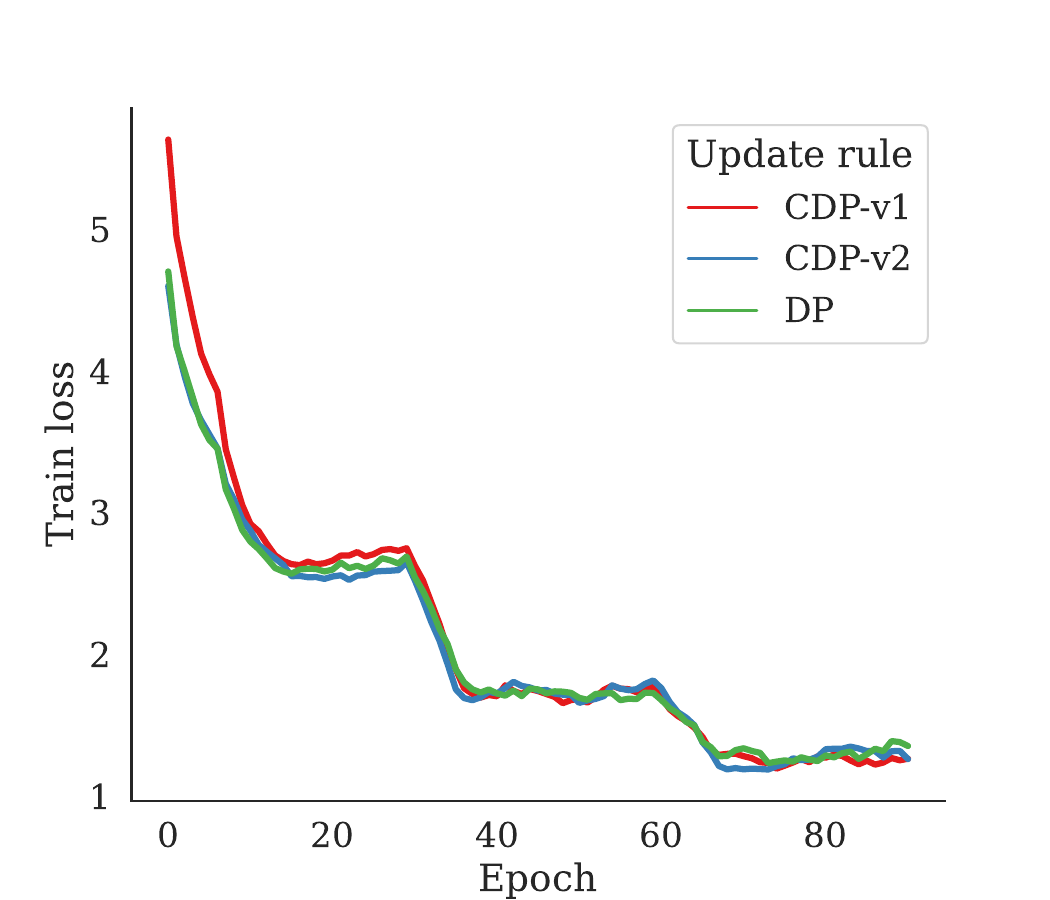}
    \caption{\textbf{Training loss of a ResNet-50 trained on ImageNet} following the learning rules \eqref{eq:ur_t}, \eqref{eq:ur_tm1} and \eqref{eq:ur_int}. Values are averaged over a window of 7 epochs for the sake of readability. The loss of CDP-v1 is significantly higher at the beginning of training, which is not the case for CDP-v2. As parameters converge, the effect of the delay disappears and the three losses show a similar training curve, with a small advantage for both CDP-v1 and CDP-v2. This confirms that the delay in our update rules does not affect convergence, even in realistic settings. }
    \label{fig:trainloss}
\end{figure}

\paragraph{Results.} We report in Tab. \ref{tab:xp_rules_cifar10} and Tab. \ref{tab:xp_rules_imagenet} the test accuracy of our models for the three rules CDP-v1, CDP-v2, and DP, for 5 runs on CIFAR-10 and 3 runs on ImageNet respectively. The variances reported in our runs are less than 0.08, indicating that our experiments lead to relatively stable results. Both tables show that CDP leads to similar or better performances compared to DP, which is consistent with the experimental findings of \cite{narayanan2021memory, ren2021zerooffload}. For CIFAR-10, CDP-v2 significantly outperforms CDP-v1, showing our improvement over PipeDream-2BW's rule. % the improvement of our learning rule over the one of PipeDream-2BW. 

We also provide in Fig. \ref{fig:trainloss} the training loss of the three concurrent learning rules on the ImageNet dataset: we note that the final loss of DP is slightly higher than both CDP-v1 and CDP-v2, with similar generalization performance, which indicates how close the three methods are, both from an optimization and statistical point of view. This is consistent with the theoretical insights of \cite{mishchenko2023asynchronous} (see Sec. \ref{sec:3-cdp}).

\begin{figure}
\centering
    \includegraphics[width=0.8\linewidth]{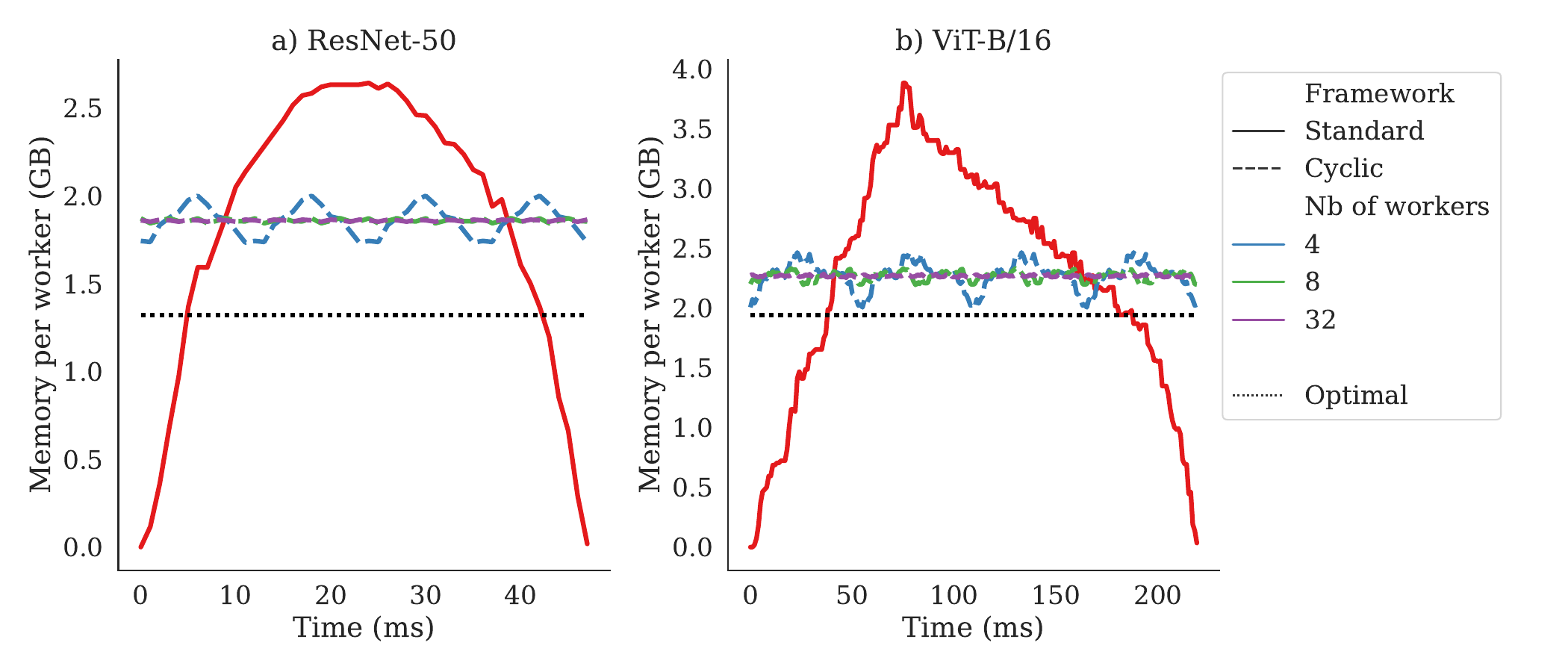} 
    \caption{\textbf{Activation memory per worker, when training with $N$ workers on ImageNet} with an efficient implementation of DP (full) and a CDP (dashed), on a ResNet-50 and a ViT-B/16. An optimal halving of the parameters is represented in black ('Optimal'). The memory required by a forward-backward pass for one work is first tracked, and parameter memory is removed. The figure is extrapolated by mimicking the total memory used by $N$ workers training on DP (\ie simultaneously) or CDP (\ie cyclically), and dividing by $N$. As $N$ increases, the memory required by CDP flattens, to a value lower than DP. This value reaches close to the reduction in half theorized for the ViT-B/16, with $42\%$. The heterogeneity of the layers of the ResNet reduces the effectiveness of CDP, only reaching $30\%$ reduction.}
    \label{fig:memory}
\end{figure}

\paragraph{Activation memory tracking.} In Fig. \ref{fig:memory}, we track the memory used during one forward-backward pass of a ResNet-50 and a ViT-B/16 training on ImageNet. By substracting the constant value corresponding to the parameter memory, we obtain the activation memory. From this memory curve, which clearly shows the activation memory peaking at the end of the forward pass, we extrapolate for $N=4$, $8$, and $32$ the activation memory usage per worker (total memory across $N$ homogeneous workers divided by $N$) that an efficient implementation of DP and CDP would use. As expected, the activation memory used varies less during training with CDP, flattening as $N$ increases.  
Furthermore, the maximum activation memory used by the DP paradigm is significantly higher than that of CDP. The activation memory ratio we find here is approximately $30\%$ for the ResNet-50 and $42\%$ ($=\frac{3.9-2.3 \text{ GB}}{3.9\text{ GB}}$) for the ViT-B/16, close to the ideal halving theorized. The ResNet reaches a lower memory improvement due to the heterogeneity of its layers, which require varying activation memory for the same execution time (as feature size decreases through depth). This is not an issue for a Transformer-based model, as feature size stays constant across depth. This confirms that CDP will significantly improve the memory used in real implementations when using homogeneous stages and workers.

\section{Conclusion}

We introduced Cyclic Data Parallelism (CDP), an alternative framework to Data Parallelism. By executing the forward and backward passes of micro-batches of data cyclically rather than simultaneously, we balance gradient communications and the total memory occupied by activations.
We particularize CDP within the context of Data, Model, and Pipeline Parallelism as well as ZeRO-DP, and demonstrate improvements in the total memory required to store activations or in the number of communication steps required between time steps during training.
In particular, CDP reduces the number of devices needed in MP and reduces the communication delay in ZeRO-DP. 
Our results are supported by existing theoretical guarantees in the small delay settings.
Finally, our numerical experiments on ImageNet show our update rules achieve similar testing accuracy as standard DP. 

In future work, we would like to release a highly efficient implementation compatible with cuDNN frameworks, as well as further relax our update rule to explore the possibility of using asynchronous and random delays. 

\section*{Acknowledgements}

This work was supported by Project ANR-21-CE23-0030 ADONIS, EMERG-ADONIS from Alliance SU, and Sorbonne Center for Artificial Intelligence (SCAI) of Sorbonne University (IDEX SUPER 11-IDEX-0004). This work was granted access to the AI resources of IDRIS under the allocations 2022-AD011013095 and 2023-A0151014526 made by GENCI. 

\bibliographystyle{splncs04}
\bibliography{main}
\end{document}